\documentclass[conference, 11pt]{IEEEtran} 
\IEEEoverridecommandlockouts

\usepackage[letterpaper, top=1in, bottom=1in, left=1in, right=1in]{geometry}
\usepackage{xcolor}
\usepackage[english]{babel}  
\usepackage[font=footnotesize, labelfont=bf]{caption}  

\usepackage{setspace}
\usepackage{array}
\usepackage{cite}
\usepackage{amsmath, amssymb, amsfonts}
\usepackage{algorithmic}
\usepackage{graphicx}
\usepackage{subcaption}
\usepackage{booktabs}
\usepackage{tabularx}
\usepackage{multirow}
\usepackage{adjustbox}
\usepackage{url}
\usepackage{microtype}
\usepackage{fancyhdr}

\usepackage{url}

\usepackage{microtype}

\def\BibTeX{{\rm B\kern-.05em{\sc i\kern-.025em b}\kern-.08em
    T\kern-.1667em\lower.7ex\hbox{E}\kern-.125emX}}
\begin{document}

\title{VideoSAM: A Large Vision Foundation Model for High-Speed Video Segmentation\\

}

\author{
    Chika Maduabuchi, Ericmoore Jossou, Matteo Bucci \\
    \textit{Massachusetts Institute of Technology, USA} \\
    \{chika691, ejossou, mbucci\}@mit.edu
}

\maketitle

\begin{abstract}
High-speed video (HSV) segmentation is essential for analyzing dynamic physical processes in scientific and industrial applications, such as boiling heat transfer. Existing models like U-Net struggle with generalization and accurately segmenting complex bubble formations. We present VideoSAM, a specialized adaptation of the Segment Anything Model (SAM), fine-tuned on a diverse HSV dataset for phase detection. Through diverse experiments, VideoSAM demonstrates superior performance across four fluid environments—Water, FC-72, Nitrogen, and Argon—significantly outperforming U-Net in complex segmentation tasks. In addition to introducing VideoSAM, we contribute an open-source HSV segmentation dataset designed for phase detection, enabling future research in this domain. Our findings underscore VideoSAM’s potential to set new standards in robust and accurate HSV segmentation. The code and dataset used in this study are available at: \url{https://github.com/chikap421/videosam}.
\end{abstract}


\section{Introduction}
\label{sec:intro}

High-speed video (HSV) segmentation is essential for analyzing complex physical processes that are crucial in various scientific and industrial applications, such as chemical processes \cite{rajan_decimer-segmentation_2021,BRZOZOWSKI2022108003,zhang_high-speed_2020,eppel_computer_2020}, bubble recognition in heat transfer \cite{SEONG2023104336,RAVICHANDRAN2023110879,PASSONI2024104871,zhang_unifying_2023}, and high-speed imaging of dynamic events \cite{zhou2024dvisdaq,e24070942,electronics12132890,BALACHANDRAN2022104443,GAO2022396}. Manual segmentation of objects in HSVs, such as bubbles, is time-consuming, labor-intensive, and often subjective, requiring significant expertise \cite{SUH2024100309,ZHANG2019182,CHEN2023519,MALAKHOV2023104402}. Automated segmentation methods can significantly reduce the time and labor required, increase consistency, and enable the analysis of large-scale HSV datasets \cite{ronneberger2015unet,10385990,Fu_2020,RAYED2024101504,RIZWANIHAQUE2020100297}.

Convolutional neural networks (CNNs), particularly U-Net \cite{ronneberger2015unet}, have become the standard for HSV segmentation tasks \cite{RAVICHANDRAN2023110879, PASSONI2024104871, SEONG2023104336, MALAKHOV2023104402} due to their ability to learn complex features and deliver accurate segmentation results. However, these models are often highly task-specific, limiting their generalization to new tasks or varying imaging conditions \cite{he2018mask,minaee2020image,xiong2024efficient,zuo2024robust,Mahbod_2024}. This limitation is particularly challenging in HSV analysis, where dynamic nature, temporal dependencies, and real-time processing demands are prevalent \cite{minaee2020image,app11198802,Zeineldin_2020,zhou2018semantic,xiao2018unified}. 

Recent advancements in segmentation models, particularly the emergence of foundation models like the Segment Anything Model (SAM) \cite{kirillov2023segment}, have shown remarkable versatility and generalization across various tasks. Despite their success, the applicability of these models to scientific HSV tasks, especially for tasks like bubble segmentation, remains largely unexplored due to the significant differences between natural images and HSV frames \cite{hassan2023bubbleml,RAMASWAMY20024761,10.1117/1.OE.57.12.124105,thoroddsen_high-speed_2008,duan_synchronized_2013}.

Moreover, the diverse distribution of training data from different scientific HSV experiments often leads to domain shifts, which can result in suboptimal performance when using traditional CNNs for segmentation \cite{s24113363,lin2022mitigating,park2020calibrated,yan2019domain,JU2022110861,SEONG2023104336}. To address these issues, specialized models are typically built for each data distribution, which may improve performance on similar test data but often fails to generalize well to new datasets \cite{park2020calibrated,lin2022mitigating,NIPS2006_b1b0432c,Zhou_2022,cho2023complementary}. Despite the dominance of models like U-Net in HSV segmentation, there is a significant gap in leveraging large vision foundation models like SAM to achieve better generalization in these tasks.

To overcome these limitations, we introduce \textbf{VideoSAM}, a refined vision foundation model that enhances SAM's segmentation performance specifically for scientific HSV tasks. VideoSAM is fine-tuned on a newly curated and extensive dataset of HSV frame-mask pairs, designed to cover a broad range of boiling modalities and dynamic behaviors. This dataset, which we introduce as a key contribution, facilitates the fine-tuning of large vision models for more robust and generalizable segmentation in HSV analysis. 

Our experiments across four different fluids—Water, 3M\textsuperscript{TM} Fluorinert\textsuperscript{TM} Liquid (FC-72), Nitrogen, and Argon—demonstrate that VideoSAM significantly outperforms traditional models like U-Net, particularly in complex fluid environments. These results highlight the potential of VideoSAM to serve as a versatile and robust solution for HSV segmentation, offering improved generalization and accuracy in diverse scientific applications.

The remainder of this paper is organized as follows: Section 2 provides an overview of the related work in foundation models and HSV segmentation. Section 3 describes the methodology, including the dataset preparation, VideoSAM model, and training process. Section 4 presents the experimental results and discusses the performance of VideoSAM in comparison to existing models. Finally, Section 5 concludes the paper and outlines future research directions.

\section{Related Works}
\label{sec:related_works}

Deep learning-based methods have shown significant promise in scientific HSV segmentation tasks, leveraging the hierarchical feature learning capabilities of deep neural networks. This section is divided into three parts: traditional HSV segmentation methods, large vision foundation models in segmentation, and the application of these models to scientific HSV tasks.

\subsection{Traditional HSV Segmentation Methods}

CNNs have become the dominant models for bubble segmentation in HSVs. U-Net \cite{ronneberger2015unet} and Mask R-CNN \cite{he2018mask} are particularly popular due to their robust performance in various image segmentation tasks. For instance, Passoni et al. \cite{PASSONI2024104871} and Dunlap et al. \cite{dunlap2024bubbleid} effectively employed U-Net and Mask R-CNN-based models for bubble segmentation, demonstrating their capacity to capture the complex dynamics of bubbles in HSVs. Similarly, Malakhov et al. \cite{MALAKHOV2023104402} utilized a modified U-Net/Mask R-CNN architecture to segment bubbles in boiling experiments. Other researchers, such as Chernyavskiy et al. \cite{chernyavskiy_cnn-based_2021} and Hessenkemper et al. \cite{hessenkemper_bubble_2022}, explored various CNN variants for similar tasks, further validating the efficacy of CNNs in HSV segmentation.

Despite the advancements brought by these CNN architectures, they are inherently task-specific and often struggle to generalize to new HSV conditions, fluid properties, or imaging setups \cite{he2018mask,minaee2020image,xiong2024efficient,zuo2024robust,Mahbod_2024}. The diverse distribution of training data from various scientific HSV experiments can lead to domain shifts, resulting in suboptimal performance when these models are applied to unseen data distributions \cite{s24113363, lin2022mitigating, park2020calibrated, yan2019domain, JU2022110861, SEONG2023104336}. Although specialized models tailored to specific data distributions can improve performance on similar test datasets, they often fail to generalize well beyond the trained distribution \cite{park2020calibrated, lin2022mitigating, NIPS2006_b1b0432c, Zhou_2022, cho2023complementary}.

\subsection{Large Vision Foundation Models in Segmentation}

Recent advancements in image segmentation have introduced foundation models, such as the SAM \cite{kirillov2023segment}, which are pretrained on large-scale, diverse datasets and have demonstrated strong performance across various segmentation tasks. These models are designed to be versatile and generalizable, capable of adapting to different segmentation challenges with minimal fine-tuning. SAM, for instance, leverages a novel prompt-based approach to achieve state-of-the-art results on natural image segmentation benchmarks, highlighting the potential of large vision foundation models to surpass traditional CNN-based approaches in flexibility and performance.

Other notable foundation models include SEEM \cite{zou2023segment}, Mask2Former \cite{cheng2022maskedattention}, HRNet \cite{wang2020deep}, and Swin Transformer \cite{liu2021swin}. These models have set new benchmarks in natural image segmentation by incorporating architectural innovations such as multi-modal prompts, masked attention mechanisms, high-resolution representations, and hierarchical feature extraction. Their success in natural image segmentation has spurred applications in various domains, including medical imaging \cite{Maquiling_2024, Huang_2024, zhang2024segment, Ma_2024, Mazurowski_2023}, remote sensing \cite{OSCO2023103540, YANG2024103929, 10522788, Ding_2024, ji2023segment}, and video tracking \cite{yang2021associating, cheng2023segment, maalouf2024follow, yang2023track, zhou2024sampd}.

\subsection{Existing Datasets in Boiling Phenomena}
Existing datasets, such as the Boiling Dataset \cite{Dunlap2022SupervisedAU}, focus primarily on broader boiling phenomena for classification tasks. However, these datasets do not cover phase detection data segmentation using HSV, lacking the detailed, high-resolution frame-mask pairs necessary for training and fine-tuning advanced large vision models in HSV research. We introduce a novel dataset specifically designed for phase detection in HSV segmentation, addressing these limitations and pushing forward the application of large vision models in boiling data analysis.

\subsection{Application of Large Vision Models to Scientific HSV Segmentation}

The application of large vision foundation models like SAM to scientific HSV segmentation remains largely unexplored, despite their potential to address the limitations of traditional CNN models in this domain. The unique challenges posed by HSVs, such as overlapping bubbles, weak boundaries, and dynamic bubble behavior, require more sophisticated segmentation approaches than what traditional CNN models can offer. Large vision models, with their ability to learn from diverse data and generalize across tasks, present an opportunity to significantly advance HSV segmentation.

However, the gap in applying these models to HSV segmentation is notable. Traditional CNN models dominate this field, yet they often fail to generalize well to new HSV conditions or fluid properties. The lack of open-source datasets specific to phase detection in HSVs has further hindered the development and fine-tuning of large vision models in this area. To bridge this gap, our work introduces \textbf{VideoSAM}, a refined segmentation foundation model that enhances SAM for HSV tasks. VideoSAM is fine-tuned on a specialized dataset of HSV frame-mask pairs, incorporating domain-specific adaptations that enable the model to capture the unique characteristics of bubbles effectively. 

In addition to introducing VideoSAM, we contribute a novel HSV segmentation dataset specifically designed for phase detection. This dataset is made publicly available to facilitate further research and development of large vision foundation models for HSV segmentation. Through comprehensive experiments, we demonstrate that VideoSAM outperforms both SAM and specialist models like U-Net across various HSV datasets, making it a promising candidate for advancing segmentation tasks in scientific and industrial settings.

\section{Methodology}
\label{sec:methodology}

\subsection{Model Architecture}
\label{subsec:model_architecture}

\begin{figure}[htb]
  \centering
  \includegraphics[width=0.5\textwidth]{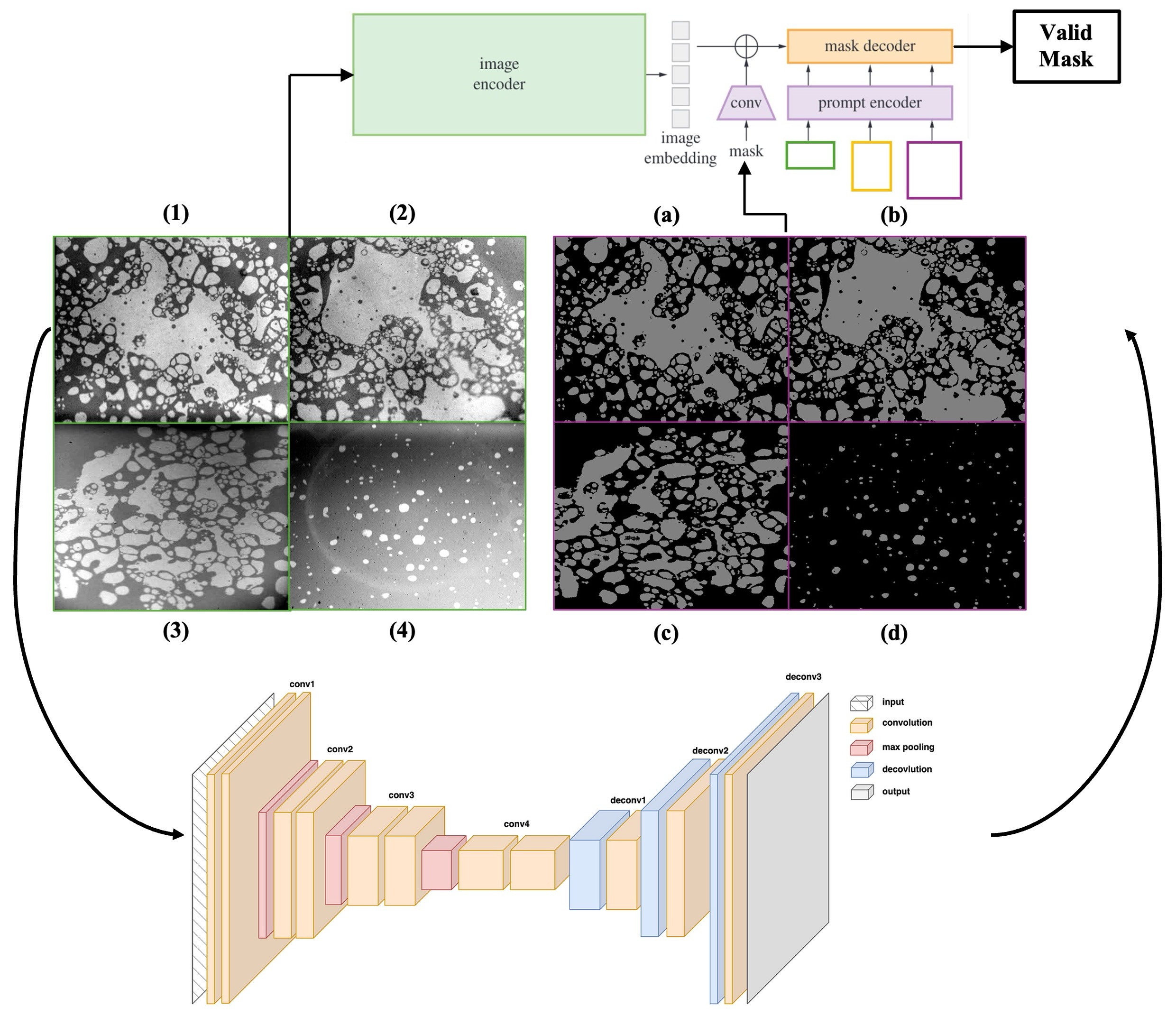}
  \caption{Illustration of the VideoSAM model architecture and integration with U-Net CNN. The initial segmentation masks generated by fine-tuned U-Net models for each modality are paired with their respective images and fed into the VideoSAM transformer. The image encoder and mask decoder process these inputs to refine the segmentation, leveraging the pre-trained SAM components for HSV segmentation.}
  \label{fig:videosam_model}
\end{figure}

The model architecture, illustrated in Figure \ref{fig:videosam_model}, follows a two-stage approach. Initially, specialized U-Net CNNs were built for each data modality (Argon, Nitrogen, FC-72, and Water) since the model could not generalize across all modalities. These U-Net models, originally trained on cellular images \cite{ronneberger2015unet}, were fine-tuned to each modality to generate initial segmentation masks. The image-mask pairs were then fed into the VideoSAM transformer architecture to test its zero-shot generalization ability across different fluids. The model processes these inputs via its image encoder and mask decoder, producing refined segmentation masks for the final output. This approach leverages U-Net’s initial mask generation and VideoSAM’s capacity for handling complex high-speed video segmentation tasks.

\begin{table}[htb]
  \caption{High-Speed Video Modalities for Experimental Data Collection}
  \label{tab:video_modalities}
  \centering
  \begin{tabularx}{\columnwidth}{@{}l l c c@{}}
    \toprule
    \textbf{Modality} & \textbf{Conditions} & \textbf{Heat Flux} & \textbf{Frames} \\
    & & (kW/m\(^2\)) & \\
    \midrule
    Argon     & SPB                      & 120  & 6K \\
    Nitrogen  & SPB                      & 120  & 6K \\
    FC-72     & SPB                      & 170  & 6K \\
    Water     & FB (500 kg/m\(^2\)s)     & 3000 & 7.5K \\
    \bottomrule
  \end{tabularx}
  \vspace{0.2cm}
  \footnotesize{SPB: Saturated Pool Boiling, FB: Flow Boiling at 500 kg/m\(^2\)s}
\end{table}

\subsection{Data Collection}
\label{subsec:data_collection}

To develop a versatile and generalizable foundation model for bubble segmentation in HSVs, we curated an extensive and diverse dataset of video frame-mask pairs from high-speed camera experiments. The dataset encompasses several boiling modalities with varied conditions, fluid properties, pressures, and imaging setups, ensuring robustness and applicability across a wide range of scenarios. The HSV modalities are detailed in Table \ref{tab:video_modalities}, which includes data collected under saturated pool boiling (SPB) conditions at 1 bar for liquid argon, nitrogen, and FC-72. Additionally, high-pressure flow boiling (FB) experiments were conducted for water at 10 bar with a mass flux of 500 kg/m\(^2\)s. Each modality was captured with specific resolutions and frame counts, effectively recording the dynamic behavior of these boiling processes.

\subsection{Data Processing}
\label{subsec:data_processing}

To ensure high-quality training data, 250 random frames were sampled from each data modality, yielding 1000 frames in total. Although this study focuses on frame-based segmentation, future work will incorporate temporal dynamics to enhance performance in HSV tasks. The training, testing, and validation sets were created using an 80:20 split of the remaining frames, ensuring diverse and representative samples across all modalities.

Raw images were converted to grayscale and normalized to enhance feature visibility by subtracting blank reference frames and adjusting contrast, which reduced background noise and improved mask extraction. Ground truth segmentation masks were created using a combination of manual annotation and semi-automatic techniques, with ImageJ \cite{rueden2017imagej2} and adaptive thresholding algorithms \cite{ZHANG2019182} used by domain experts. A U-Net model was trained on these annotated frames to generate initial segmentation masks, which were then refined by human annotators to scale the dataset efficiently while maintaining accuracy.

\begin{figure}[tb]
  \centering
  \includegraphics[width=0.45\textwidth]{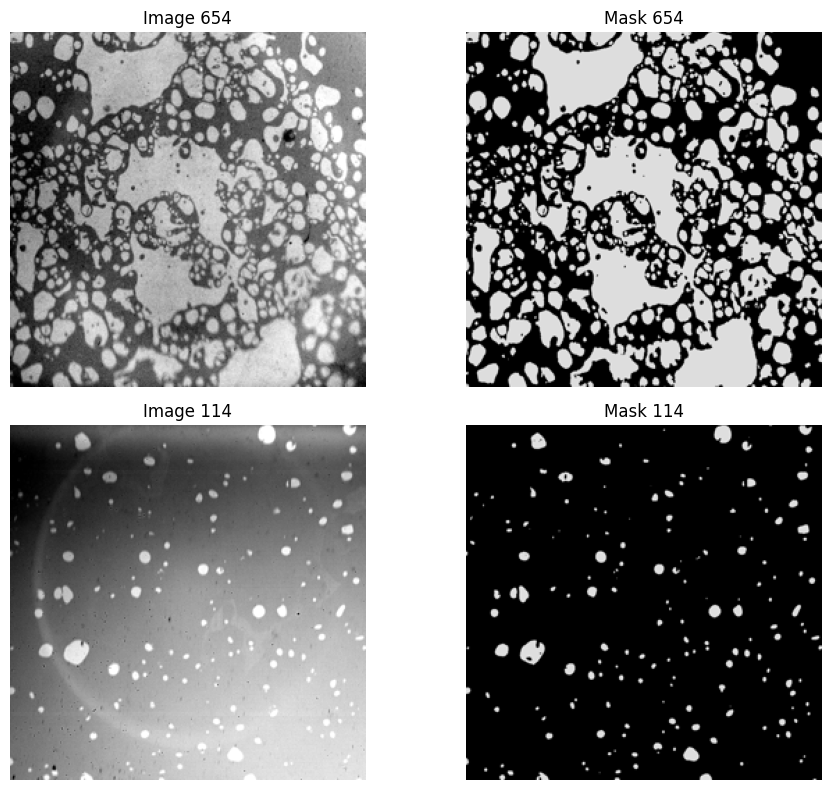}
  \includegraphics[width=0.45\textwidth]{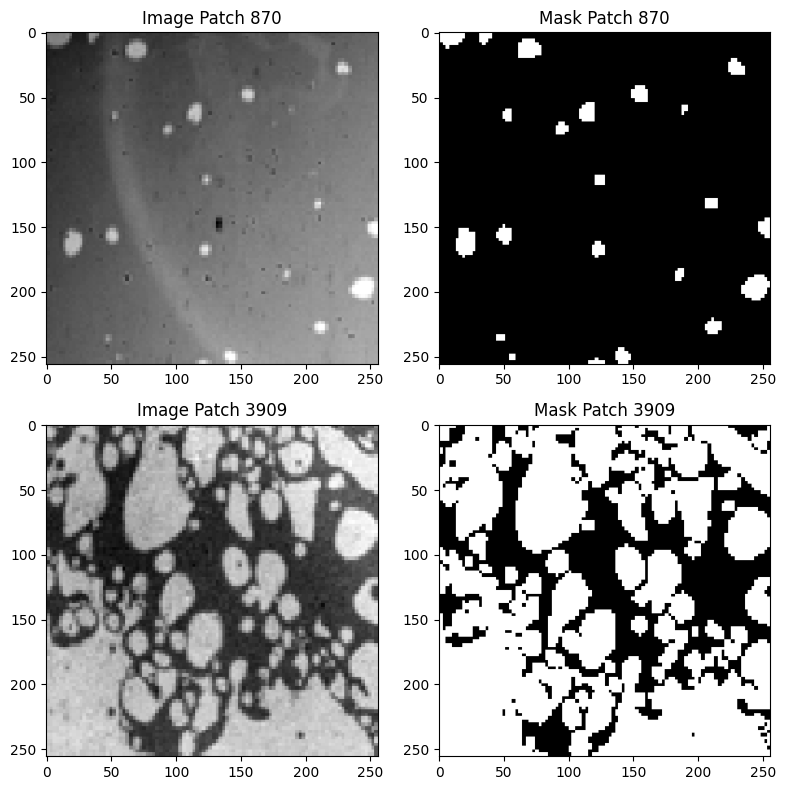}
  \caption{Left: Original high-speed video frames showcasing randomly sampled frames from the large training dataset. The images illustrate the difference between the modalities of water (image 114) and gas (image 654). Notice the difference in bubble footprints, with gases exhibiting more bubbles with complex shapes compared to water. Right: Visualization of the patched images resulting from the patchification process. This process highlights the segmentation of the original images into smaller patches for detailed analysis.}
  \label{fig:patchification}
\end{figure}

The images and masks were then patchified using a 100x100 pixel grid, discarding patches without mask information. Remaining patches were resized to 256x256 pixels for model training, and masks were normalized to binary values (0 or 1). Random patches were visualized to confirm preprocessing integrity, as shown in Figure \ref{fig:patchification}.

This comprehensive data processing pipeline ensured the quality, diversity, and representativeness of the training data, improving the robustness and generalizability of the VideoSAM model.

\subsection{Training Process}
\label{subsec:training_process}

VideoSAM was fine-tuned by freezing the pre-trained vision and prompt encoder layers of the \texttt{facebook/sam-vit-base} model while allowing updates to the mask decoder. A custom \texttt{SAMDataset} class managed the data, incorporating bounding box generation for masks. The dataset was wrapped in \texttt{DataLoader} objects to enable efficient batch processing.

The model was trained using the Adam optimizer (\(1 \times 10^{-5}\), no weight decay) with a combination of Dice Coefficient and Cross-Entropy Loss. Mixed precision training with \texttt{GradScaler} was applied to accelerate training and reduce memory usage, with gradient clipping used to maintain stability. At each epoch, the model's performance was evaluated on a validation set using metrics like IoU, precision, and recall. Learning rate scheduling (\texttt{ReduceLROnPlateau}) was applied based on validation loss to optimize training.

Training and validation losses, as well as key metrics, were logged, and model checkpoints were saved periodically. This meticulous process ensured that VideoSAM was robust and capable of performing high-accuracy bubble segmentation across diverse data modalities.

\subsection{Inference Pipeline}

The inference pipeline for VideoSAM evaluated performance on different data modalities using mask extraction and metrics evaluation for both single and composite frames. 

Mask extraction began by converting high-speed video frames to grayscale and normalizing them. For single frames, the \texttt{SAMInferenceDataset} class segmented images into smaller patches using grid-based bounding boxes. These patches were processed through both VideoSAM and the base SAM model, and the predicted masks were stitched together to reconstruct full image masks. For composite frames, the same approach was extended to sequences, ensuring dynamic video data was consistently evaluated over time.

Metrics evaluation compared predicted masks with ground truth using F1 Score, IoU, and Precision for single frames. For composite frames, these metrics were aggregated across sequences to provide detailed insights into performance consistency, including mean, minimum, maximum, and standard deviation.

This pipeline combined mask extraction and detailed metrics to thoroughly assess VideoSAM’s segmentation capabilities across diverse modalities and ensure robust evaluation.

\subsection{Experimental Setup}

In this study, we conducted three key experiments to evaluate the performance of VideoSAM. These experiments were designed to test the model’s zero-shot/generalization capabilities, its performance across different data modalities, and to compare its results against a diversified CNN model.

\subsubsection{Experiment 1: Zero-Shot Generalization Across Modalities}
The first experiment aimed to test the ability of VideoSAM to generalize to unseen data modalities. For this, we trained VideoSAM on the high-speed video frames from the Argon boiling modality. After training, we tested the model on all other data modalities, including Nitrogen, FC-72, and Water. The goal was to evaluate how well the model could perform zero-shot segmentation on data modalities it had not encountered during training. We inspected the results through both visual analysis and by quantifying key metrics, including Intersection over Union (IoU) and F1 Score, which are standard metrics for segmentation tasks. We expected VideoSAM to outperform the baseline Segment Anything Model (SAM) across all modalities. The experiment demonstrated that VideoSAM achieved superior segmentation quality compared to SAM, as confirmed by both visual inspection and the quantified metrics.

\subsubsection{Experiment 2: Performance Across Multiple Modalities}
In the second experiment, we evaluated VideoSAM's ability to handle multiple boiling modalities. The model was trained on a combination of four different datasets, representing boiling processes in Argon, Nitrogen, FC-72, and Water. After training, the model was tested on unseen data from these same modalities to determine how well it could generalize across diverse conditions. This experiment was crucial for assessing the robustness of VideoSAM in handling a variety of fluids with different boiling characteristics. We expected VideoSAM to demonstrate high performance across all datasets, consistently outperforming SAM in both IoU and F1 Score. As anticipated, VideoSAM excelled across all test datasets, showing superior performance in capturing the complexities of each boiling modality.

\subsubsection{Experiment 3: Comparison with U-Net CNN}
In this experiment, we compared VideoSAM with U-Net, a well-established CNN architecture frequently employed for high-speed video segmentation tasks \cite{RAVICHANDRAN2023110879, PASSONI2024104871, SEONG2023104336}. U-Net's proven success in segmenting complex cellular structures, which share similarities with the bubble footprints in HSV data, makes it a conventional and strong baseline for comparison in these tasks. Both VideoSAM and U-Net were trained on the same four datasets—Argon, Nitrogen, FC-72, and Water—and evaluated using IoU and F1 Score. We expected VideoSAM to excel in complex fluids like FC-72, Nitrogen, and Argon, where dynamic and intricate boiling behaviors dominate, while U-Net was anticipated to perform better on simpler tasks, such as those in the Water dataset. As predicted, VideoSAM delivered superior results in the more challenging fluid environments, while U-Net slightly outperformed in the Water dataset, reinforcing its effectiveness in handling simpler segmentation tasks.

\section{Results and Discussions}
\label{sec:results}

\subsection{Experiment 1: Zero-Shot Generalization Across Modalities}

\begin{figure*}[htb]
  \centering
  \begin{subfigure}{0.9\textwidth}
    \centering
    \includegraphics[width=\textwidth]{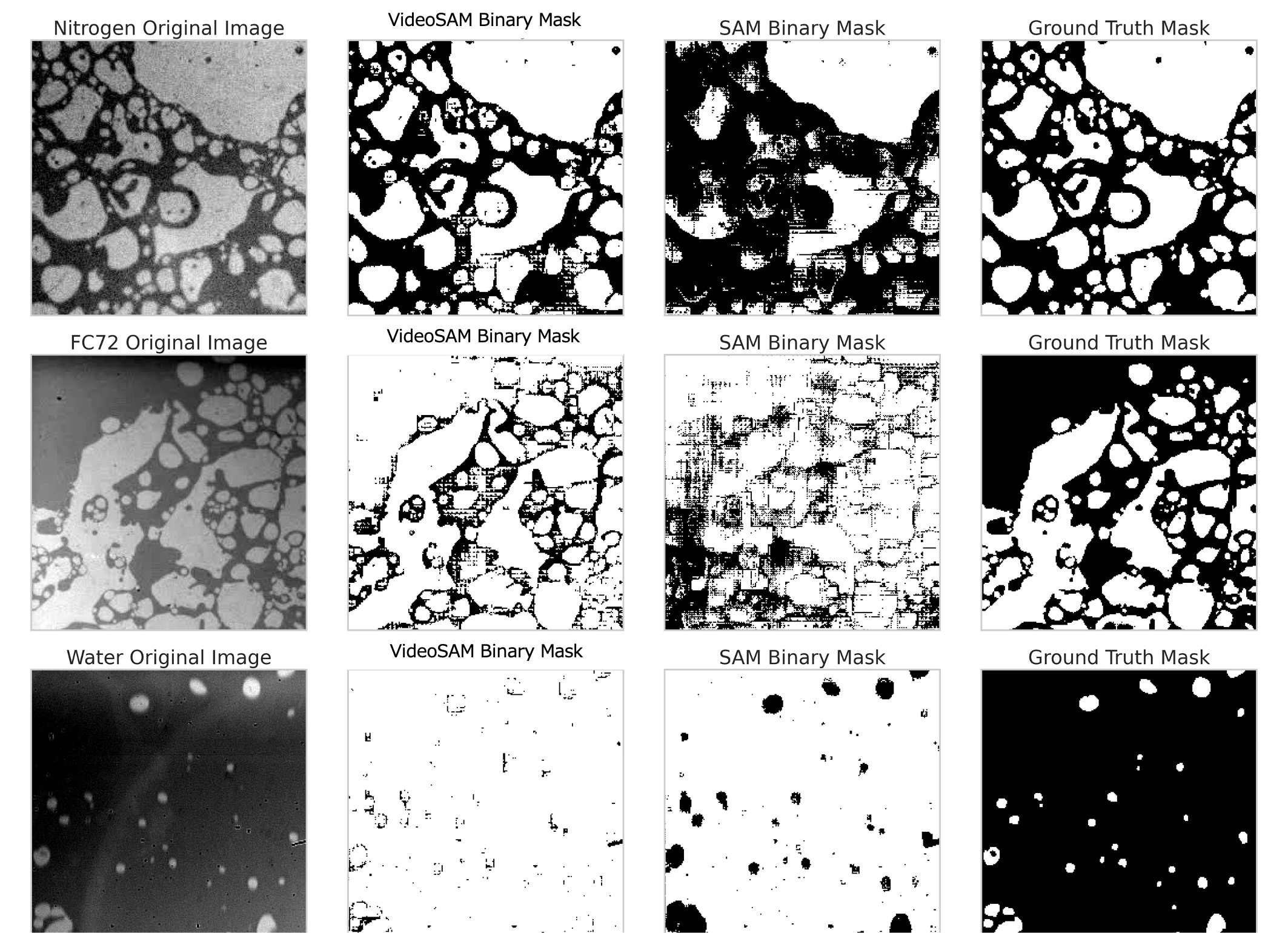}
    \caption{Comparison of Binary Masks for Nitrogen, FC-72, and Water datasets between VideoSAM, SAM, and Ground Truth Masks.}
    \label{fig:comparison_masks}
  \end{subfigure}%
  \hspace{0.5cm} 
  
  \begin{subfigure}{0.9\textwidth}
    \centering
    \includegraphics[width=\textwidth]{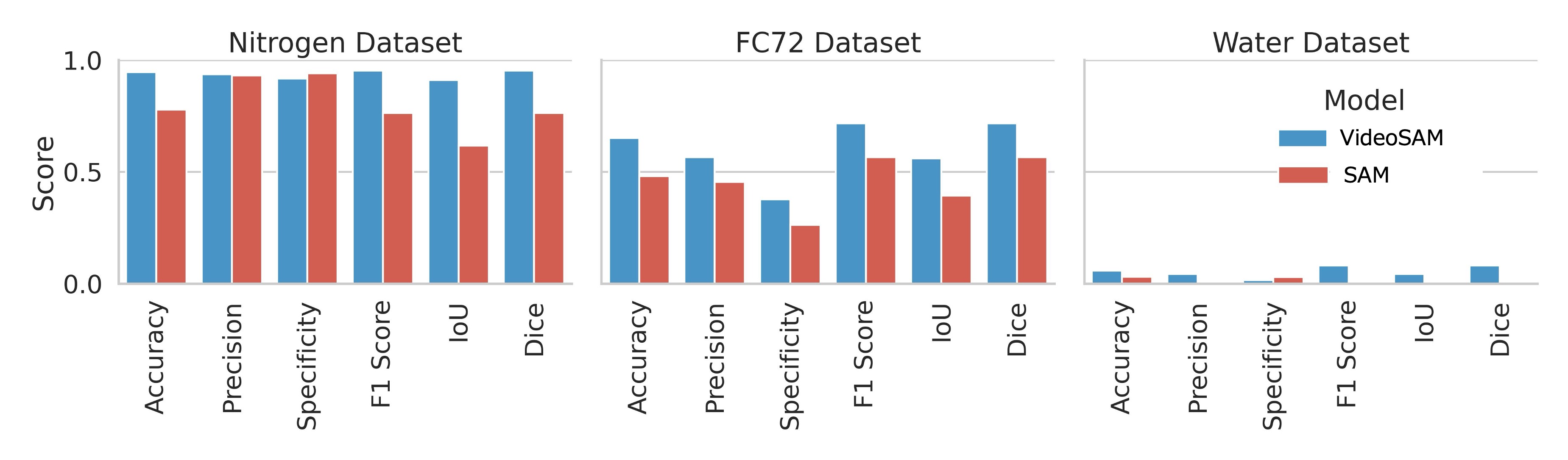}
    \caption{Performance comparison between VideoSAM and SAM across various metrics for the Nitrogen, FC-72, and Water datasets.}
    \label{fig:exp1_metrics}
  \end{subfigure}
  
  \caption{Combined results of Experiment 1: Qualitative and quantitative analysis of VideoSAM's zero-shot generalization performance.}
  \label{fig:exp1_combined}
\end{figure*}

The zero-shot performance of VideoSAM in generalizing to unseen data modalities was evaluated on Nitrogen, FC-72, and Water datasets, as shown in Figure \ref{fig:exp1_combined}. The figure combines both the qualitative analysis of segmentation masks (Figure \ref{fig:comparison_masks}) and the quantitative performance metrics (Figure \ref{fig:exp1_metrics}) to offer a comprehensive assessment of the model's zero-shot generalization capabilities.

In the \textbf{qualitative analysis} (Figure \ref{fig:comparison_masks}), VideoSAM exhibits remarkable segmentation performance in the Nitrogen and FC-72 datasets, closely matching the ground truth masks. The contours and boundaries of the bubbles are significantly better preserved in the binary masks generated by VideoSAM compared to the baseline SAM model, which struggles with boundary delineation and introduces noticeable artifacts. In particular, VideoSAM effectively captures the complex bubble structures in the Nitrogen dataset, maintaining consistent bubble segmentation even in high-density regions. Similarly, for the FC-72 dataset, VideoSAM provides a more accurate segmentation of overlapping and irregularly shaped bubbles, further reinforcing its robustness in generalizing to diverse fluid environments.

However, the model's performance degrades on the Water dataset, where fewer objects of interest and substantial background noise negatively affect its segmentation accuracy. The binary mask for the Water dataset generated by VideoSAM fails to distinguish bubbles clearly, highlighting a significant challenge in simpler datasets with low contrast and fewer distinct objects.

The \textbf{quantitative analysis} (Figure \ref{fig:exp1_metrics}) supports these observations. VideoSAM outperforms SAM across the Nitrogen and FC-72 datasets in terms of accuracy, precision, IoU, and Dice coefficient. The model achieves notably higher IoU and F1 scores, confirming its ability to generalize well to unseen data with complex bubble distributions. For example, in the Nitrogen dataset, VideoSAM shows substantial improvement in accuracy and specificity, translating to better boundary detection and fewer false positives.

Conversely, in the Water dataset, VideoSAM’s performance falls significantly below that of the other datasets, as reflected in its poor IoU and Dice scores. The simplicity of the Water dataset, characterized by fewer bubbles and more uniform backgrounds, proves to be a challenge for the model, suggesting that its architecture may be overfitted to the complexities of dense and overlapping structures, thus making it less effective in simpler scenarios.

In conclusion, while VideoSAM demonstrates impressive zero-shot generalization in datasets with intricate fluid dynamics like Nitrogen and FC-72, its performance on simpler datasets like Water reveals a limitation that may require additional architectural or preprocessing adjustments to handle different types of modalities effectively.

\subsection{Experiment 2: Performance Across Multiple Modalities}

In this experiment, the goal was to evaluate the VideoSAM model's generalization performance across various high-speed video modalities, specifically for fluids like Water, FC-72, Nitrogen, and Argon. The results were evaluated based on single-frame and composite-frame segmentation tasks. In both analyses, we compare VideoSAM with the baseline SAM model.

\begin{figure*}[htb]
    \centering
    \begin{subfigure}[b]{0.48\textwidth}
        \centering
        \begin{tabular}{lcccc}
            \toprule
            \textbf{Fluid} & \textbf{Model} & \textbf{IoU} & \textbf{F1 Score} \\
            \midrule
            \multirow{3}{*}{\textbf{Water}}  
              & U-Net     & \textbf{0.5619} & \textbf{0.7191} \\
              & SAM       & 0.0620 & 0.1165 \\
              & VideoSAM  & 0.1894 & 0.3143 \\
            \midrule
            \multirow{3}{*}{\textbf{FC-72}}  
              & U-Net     & 0.7244 & 0.8400 \\
              & SAM       & 0.5721 & 0.7278 \\
              & VideoSAM  & \textbf{0.7997} & \textbf{0.8885} \\
            \midrule
            \multirow{3}{*}{\textbf{Nitrogen}}  
              & U-Net     & 0.7547 & 0.8602 \\
              & SAM       & 0.6702 & 0.8025 \\
              & VideoSAM  & \textbf{0.8317} & \textbf{0.9080} \\
            \midrule
            \multirow{3}{*}{\textbf{Argon}}  
              & U-Net     & 0.7815 & 0.8773 \\
              & SAM       & 0.6464 & 0.7852 \\
              & VideoSAM  & \textbf{0.8384} & \textbf{0.9120} \\
            \bottomrule
        \end{tabular}
        \caption{Comparison of IoU and F1 Score for U-Net, VideoSAM, and SAM across fluids.}
        \label{tab:unet_videosam_sam_comparison}
    \end{subfigure}%
    \hspace{0.5cm} 
    
    \begin{subfigure}[b]{0.7\textwidth}
        \centering
        \includegraphics[width=\textwidth]{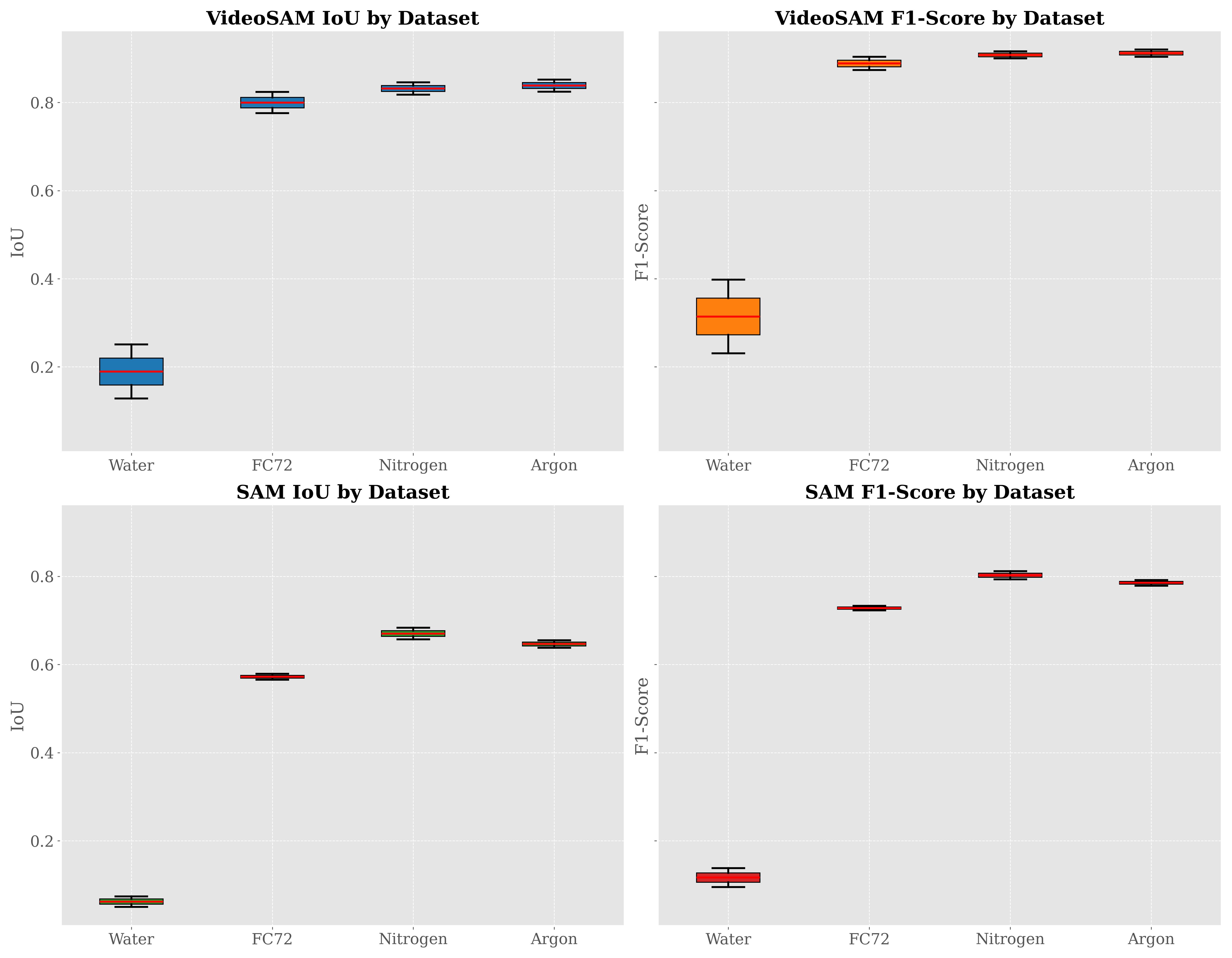}
        \caption{Box plots comparing IoU and F1 Score across datasets.}
        \label{fig:composite_frame_analysis}
    \end{subfigure}

    \caption{Combined table and figure layout comparing the performance of U-Net, VideoSAM, and SAM across different datasets.}
    \label{fig:combined_performance}
\end{figure*}

Table \ref{tab:unet_videosam_sam_comparison} presents the results of the single-frame analysis for VideoSAM across different datasets. As indicated, VideoSAM consistently outperforms SAM in the more complex datasets such as Nitrogen, FC-72, and Argon. These datasets exhibit complex bubble dynamics and intricate boundaries, and VideoSAM's architecture proves more effective in capturing these challenging fluid environments. For instance, in the Nitrogen dataset, VideoSAM achieves an IoU of 0.8317 and an F1 Score of 0.9080, compared to SAM's IoU of 0.6702 and F1 Score of 0.8025. Similarly, in the FC-72 dataset, VideoSAM outperforms SAM with an IoU of 0.7997 and F1 Score of 0.8885, showing better segmentation accuracy and boundary delineation.

The Water dataset, however, presents a unique challenge for both models due to fewer objects of interest and significant background noise. VideoSAM achieves an IoU of 0.1894, which, although an improvement over SAM's IoU of 0.0620, highlights the difficulty in segmenting this simpler environment. The lack of distinct features in the Water dataset reduces the effectiveness of both models in distinguishing bubbles from the background.

Additionally, the composite-frame analysis (Figure \ref{fig:composite_frame_analysis}) further supports these findings. By aggregating the model performance across multiple frames within each dataset, VideoSAM maintains its robustness and superior accuracy in complex environments, particularly in the Nitrogen, FC-72, and Argon datasets. The box plots highlight that VideoSAM's IoU and F1 Score distributions have higher median values across these datasets, showcasing more reliable and consistent performance than SAM. Nevertheless, the results on the Water dataset show a wider variance and lower median values, indicating more significant performance fluctuations due to the simplicity of the scene and background noise.

Experiment 2 demonstrates that VideoSAM excels in handling complex fluid dynamics with intricate bubble boundaries and overlapping structures, especially in datasets like FC-72, Nitrogen, and Argon. However, simpler environments, like the Water dataset, expose the model's limitations, particularly when faced with fewer distinct objects and more background noise. These results suggest that VideoSAM is a powerful tool for high-speed video segmentation in complex environments, but further refinement is needed to improve its performance in simpler scenes. Future directions could involve exploring techniques such as domain adaptation, multi-scale feature extraction, or data augmentation to address these limitations.

\subsection{Experiment 3: Comparison with U-Net CNN}

This experiment compares the performance of VideoSAM, U-Net, and SAM models across four fluid datasets: Water, FC-72, Nitrogen, and Argon. The primary objective was to assess how these models perform in both complex and simpler fluid environments. The metrics used for comparison are IoU and F1 Score, both of which measure segmentation accuracy.

Table \ref{tab:unet_videosam_sam_comparison} summarizes the IoU and F1 Scores for U-Net, VideoSAM, and SAM models across all datasets. The results clearly demonstrate VideoSAM’s superior performance in more complex environments, such as FC-72, Nitrogen, and Argon. For example, in the Argon dataset, VideoSAM achieved a mean IoU of 0.8384 and an F1 Score of 0.9120, outperforming both U-Net and SAM. Similarly, VideoSAM’s performance on the Nitrogen dataset (IoU: 0.8317, F1: 0.9080) also exceeded that of U-Net and SAM.

In contrast, U-Net showed the best performance in the simpler Water dataset, achieving an IoU of 0.5619 and an F1 Score of 0.7191, while VideoSAM only managed an IoU of 0.1894 and an F1 Score of 0.3143. This can be attributed to U-Net’s architecture, which has been fine-tuned for cellular-level data segmentation, making it more effective in simpler environments where the fluid characteristics closely resemble its training data.

The results from Table \ref{tab:unet_videosam_sam_comparison} indicate that while VideoSAM excels in handling more complex fluids, U-Net remains a strong contender in simpler datasets like Water. This suggests that the architecture of U-Net, particularly its cellular-level segmentation capabilities, makes it highly effective in environments with fewer objects and more consistent fluid structures, as seen in the Water dataset.

On the other hand, VideoSAM’s ability to handle complex fluid dynamics and intricate bubble boundaries gives it a distinct advantage in more challenging datasets like FC-72, Nitrogen, and Argon. The consistent superiority of VideoSAM in these datasets—demonstrated by its higher IoU and F1 Scores—validates its architecture’s robustness in segmenting intricate fluid behaviors and dynamic properties.

In summary, VideoSAM outperforms both U-Net and SAM in complex environments involving intricate fluid behaviors and dynamic conditions, such as FC-72, Nitrogen, and Argon. However, in simpler fluid environments like Water, U-Net’s architecture shines, surpassing VideoSAM in both IoU and F1 Score. These findings suggest that VideoSAM is particularly well-suited for complex scientific applications, while U-Net may be more appropriate for simpler segmentation tasks.

\subsection{Weaknesses and Potential Solutions}

Despite the promising results, VideoSAM exhibits certain weaknesses, particularly in handling simpler HSV datasets like Water. The model, which was fine-tuned primarily on complex fluid dynamics, tends to overfit to these intricate scenarios, leading to reduced performance on datasets with less dynamic behavior. To address this, we propose the following solutions:

\textbf{1. Hybrid Model Architecture:} 
A potential solution is the development of a hybrid model that integrates both traditional CNN layers, such as those in U-Net, and transformer-based layers from VideoSAM. The CNN layers could effectively handle simpler, static features, while the transformer layers focus on complex, dynamic features. This hybrid approach could leverage the strengths of both architectures, enhancing the model's ability to generalize across different types of datasets.

\textbf{2. Curriculum Learning:}
Curriculum learning is another approach that could significantly improve VideoSAM’s performance. By initially training the model on simpler datasets and gradually introducing more complex scenarios, the model could build a robust understanding of basic features before tackling more challenging segmentation tasks. This gradual increase in data complexity during training could improve the model’s generalization capabilities, making it more versatile across diverse HSV datasets.

\textbf{3. Multi-Scale Feature Aggregation:} 
Incorporating multi-scale feature aggregation into VideoSAM could enhance its ability to capture features at different levels of detail. This approach allows the model to effectively segment both large, simple structures and small, intricate ones, addressing the varying bubble sizes and textures across different HSV datasets. Multi-scale aggregation could be particularly beneficial in improving segmentation performance on simpler datasets like Water, where the model currently struggles.

\section{Conclusion and Future Work}
\label{sec:conclusion}

In this work, we introduced VideoSAM, a refined large vision foundation model designed for HSV segmentation, with a focus on complex boiling phenomena. VideoSAM significantly outperformed traditional models like U-Net in challenging fluid environments such as FC-72, Nitrogen, and Argon. Despite its success in intricate tasks, VideoSAM showed limitations in simpler datasets like Water, where it tended to overfit to complex scenarios.

To address this, we propose several future enhancements. These include hybrid architectures that integrate CNN layers with transformers, curriculum learning to improve generalization across data complexities, and multi-scale feature aggregation to handle varying object sizes. Expanding real-time segmentation capabilities and incorporating temporal dependencies are also key areas for future exploration.

In addition to the model, we contribute an open-source dataset for HSV segmentation, specifically curated for phase detection in boiling experiments, which will enable and facilitate further research in this domain. This dataset, along with VideoSAM’s performance improvements, advances the field of HSV analysis by providing valuable resources for future studies. Further work will also involve exploring advanced fine-tuning techniques, such as LoRA, VPT, and SSF, to enhance VideoSAM’s generalization and adaptability across different datasets.



\end{document}